\pdfoutput=1
\documentclass[11pt]{article}

\usepackage{fullpage,times}
\usepackage{parskip}
\usepackage{graphicx}

\usepackage{tikz}
\definecolor{Graylight}{gray}{0.9}
\usepackage{comment}
\usepackage{xcolor,colortbl}
\usepackage{amsmath,amssymb} 
\usepackage{color}
\usepackage{makecell}
\usepackage{bm}
\usepackage{multirow}
\usepackage{algorithm,algorithmic}
\usepackage{wrapfig}
\newcommand{\lm}[1]{\textcolor{black}{#1}}

\usepackage{lipsum}
\makeatletter
\newcommand{\printfnsymbol}[1]{%
  \textsuperscript{\@fnsymbol{#1}}%
}
\makeatother

\title{\huge Residual Mixture of Experts}
\author{
Lemeng Wu $^{1}$ \thanks{Work done during an internship at Microsoft.},  Mengchen Liu$^{2}$, Yinpeng Chen$^{2}$, Dongdong Chen$^{2}$, Xiyang Dai$^{2}$, Lu Yuan$^{2}$ \\
$^{1}$University of Texas at Austin $^{2}$Microsoft \\
{\tt\small\ lmwu@cs.utexas.edu},
{\tt\small \{mengcliu,yiche,dochen,xidai,luyuan\}@microsoft.com }
\\
}

\date{}

\begin{document}
\maketitle

\begin{abstract}
Mixture of Experts (MoE) is able to scale up vision transformers effectively.
However, it requires prohibiting computation resources to train a large MoE transformer.
In this paper, we propose Residual Mixture of Experts (RMoE), an efficient training pipeline for MoE vision transformers on downstream tasks, such as segmentation and detection.
RMoE achieves comparable results with the upper-bound MoE training, while only introducing minor additional training cost than the lower-bound non-MoE
training pipelines.
The efficiency is supported by our key observation: the weights of an MoE transformer can be factored into an input-independent core and an input-dependent residual.
Compared with the weight core, the weight residual can be efficiently trained with much less computation resource, e.g., finetuning on the downstream data.
We show that, compared with the current MoE training pipeline, we get comparable results while saving over 30\% training cost.
When compared with state-of-the-art non-MoE transformers, such as Swin-T / CvT-13 / Swin-L, we get +1.1 / 0.9 / 1.0 mIoU gain on ADE20K segmentation and +1.4 / 1.6 / 0.6 AP gain on MS-COCO object detection task with less than 3\% additional training cost.
\end{abstract}


\section{Introduction}

Vision transformers have achieved a lot of breakthroughs recently, with the strong capability to capture a large amount of data.
Such capability enables us to pretrain a vision transformer on a large-scale upstream dataset.
The pretrained model eases the finetuning on challenging downstream tasks, with a fast convergence speed and outstanding performance.

To further enhance the model capacity of vision transformers, some works scales up the vision transformers to billions of parameters that can effectively capture huge upstream data~\cite{brown2020language,zhai2021scaling}.
These models achieve state-of-the-art performances on various downstream tasks, such as semantic segmentation and object detection.
However, directly scaling up a vision transformer by increasing model width and depth will dramatically increase the training costs.
This makes the training for large vision transformers unaffordable for most vision researchers and practitioners~\cite{zhai2021scaling}.
In addition, many computer vision downstream tasks use high-resolution images as inputs.
As a result, direct scaling up is also limited by GPU memory.

To solve the problems in model scaling up, there are recent research attempts~\cite{fedus2021switch,lepikhin2020gshard,riquelme2021scaling,shazeer2017outrageously} that introduce conditional computation and sparsity in transformers.
As one of the most representative technique, Mixture of Experts (MoE) scales up a transformer with conditionally computed experts.
In an MoE transformer, each data point is only processed by a certain number of experts.
By making the components of model conditionally computed, the training cost of MoE transformers is much less than that of non-MoE transformers at the same scale.

\begin{figure}[t]
    \centering
    \includegraphics[width=0.9\textwidth]{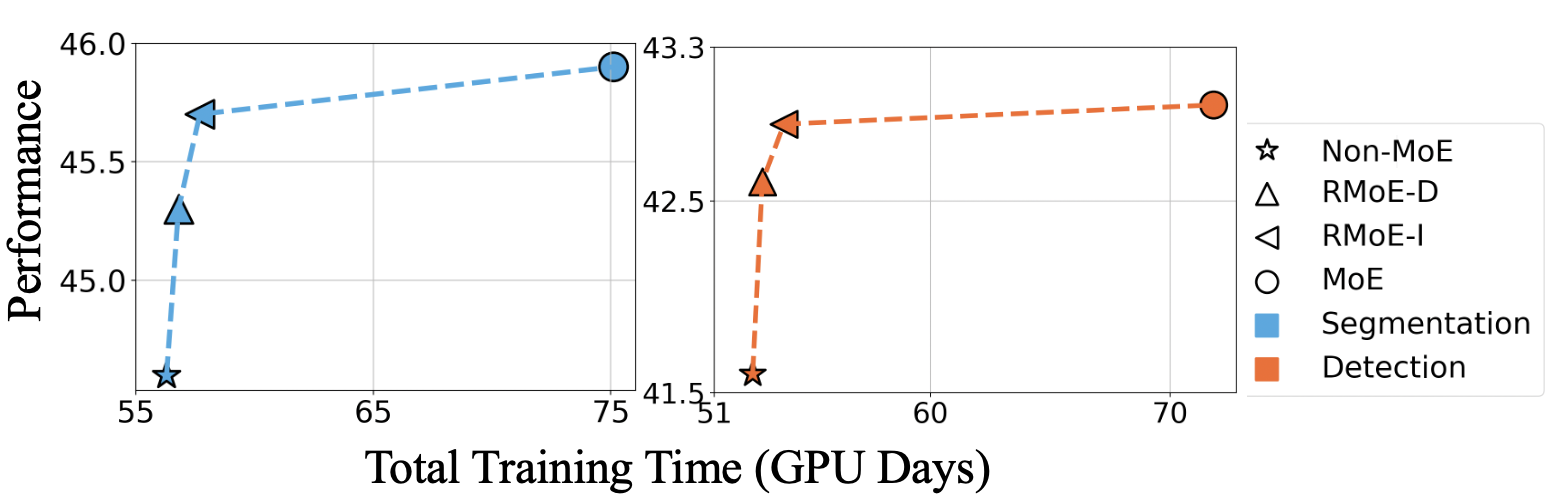}
    \vspace{-1.0em}
    \caption{RMoE: balancing performance improvement and additional training cost between upper-bound MoE and lower-bound non-MoE training pipelines. RMoE-D and RMoE-I are two variants of RMoE.}
    \vspace{-1.5em}
    \label{fig:time}
\end{figure}

\begin{figure*}[ht]
    \centering
    \includegraphics[width=0.98\textwidth]{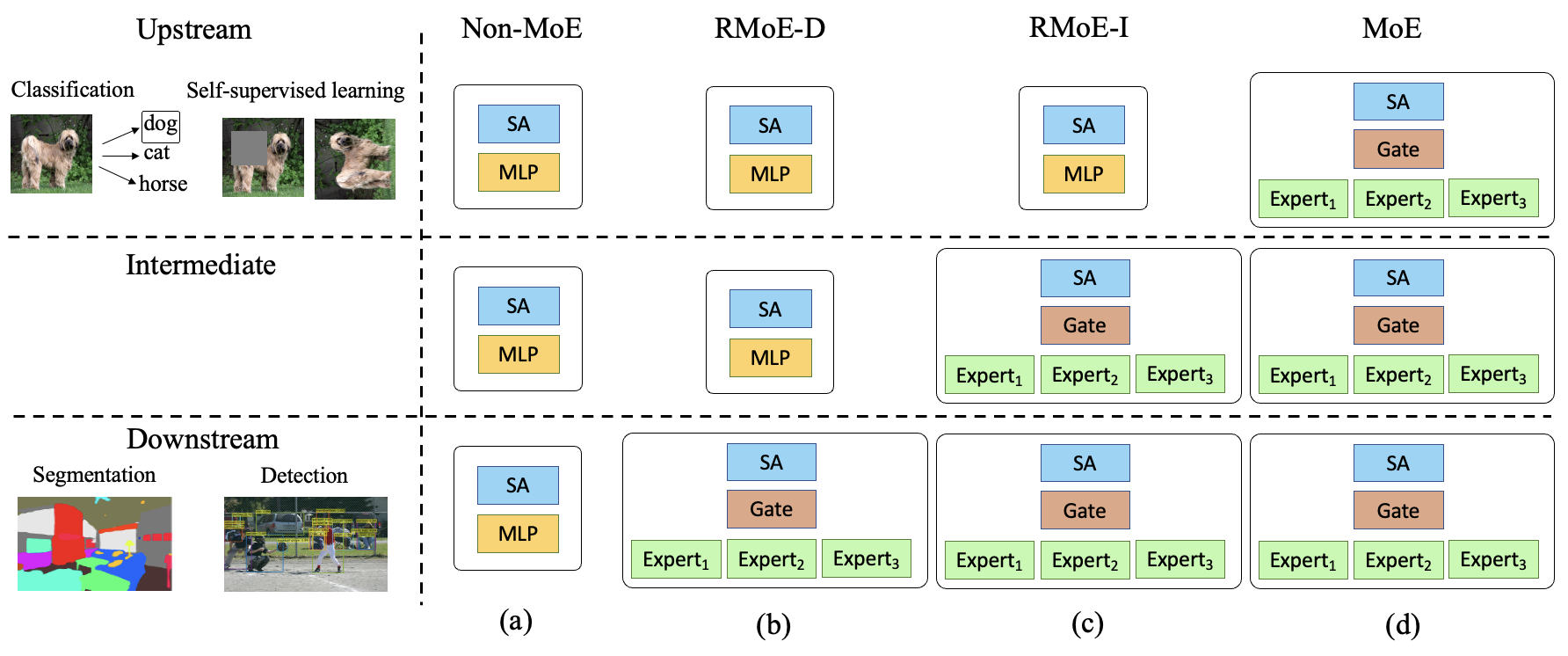}
    \vspace{-1em}
    \caption{Training pipelines for (a) non-MoE, (b)(c) RMoE and (d) MoE. Here we use 3 experts as an example to illustrate how RMoE works in the intermediate and downstream finetune stages.
    Compared with MoE training, we inherit the non-MoE transformer after the upstream pretraining to save the cost. In this figure, we simplify the general transformer block design by only showing self-attention (SA) and multilayer perceptron (MLP) modules. Norm and add operations are not shown here. }
    \label{fig:intro}
    \vspace{-1.1em}
\end{figure*}

In this work, we propose Residual Mixture of Experts (RMoE), an efficient training pipeline for large MoE vision transformers.
As shown in Fig.~\ref{fig:time}, RMoE achieves comparable results with the upper-bound MoE training pipelines, while only introducing minor additional training cost than the lower-bound non-MoE training pipelines.
The efficiency is supported by our observation from analyzing the weights of experts in a trained MoE transformer.
We find that the weights can be factored into an input-independent core and an input-dependent residual.
Although the size of the weights residual is much large than that of weights core, the residual part can be efficiently trained with much less computation resource, e.g., finetuning on the downstream data.
Using this observation, we develop the RMoE training pipeline.
As shown in Figure~\ref{fig:intro} (b) and (c), in RMoE training, we can pretrain a non-MoE transformer on the upstream data and scale up the model during downstream finetuning or intermediate finetuning.
By skipping the training of the scaled-up model on the heavy-duty upstream task, we only introduce minor additional training cost while enjoying the performance boost brought from the large model capacity.
Furthermore, since many pretrained non-MoE transformer checkpoints are publicly available, practitioners can leverage these well-trained transformers as an initialization.
Thus, our method opens the possibility to customize a large-scale transformer for various tasks without the limitation of the large amount of computation resources.
\lm{We demonstrate the effectiveness of RMoE on object detection and segmentation.
By applying RMoE to different vision backbones, Swin-T, CvT-13 and Swin-L, we get a +1.1 / 0.9 / 1.0 mIoU on ADE20K and +1.4 / 1.6 / 0.6 AP on object detection with less than 3\% additional cost.}


\section{Related Work}
\subsection{Vision Transformers}
Convolution Neural Networks (CNNs), as universal and powerful structures in computer vision domain~\cite{he2016deep,howard2017mobilenets,hu2018squeeze,huang2017densely,krizhevsky2012imagenet,sandler2018mobilenetv2,simonyan2014very,szegedy2015going,tan2019efficientnet}, get a great success in the last decade.
Recently, vision transformers, show promising results in various vision tasks and attract intense research interests.
ViT \cite{dosovitskiy2020vit}, as the first breakthrough work, models each image as a set of tokens, where each token is an image patch.
It achieves a competitive image classification result compared with traditional CNNs.
Triggered by ViT, a series of follow-up works~\cite{chu2021twins,chu2021we,dong2021cswin,el2021training,han2021tnt,he2021transreid,jiang2021token,liu2021swin,touvron2020deit,touvron2021going,wang2021pyramid,wu2021cvt,xu2021coat,yuan2021incorporating,yuan2021tokens}  continuously refresh the image classification records and push the potential of vision transformers in computer vision to a new height.
In the above works, many adopt a hierarchical architecture and a large-scale upstream data pretraining.
As a result, they unlock the ability to finetune to various downstream vision tasks like segmentation and detection.
In our work, we leverage the architectures and available checkpoints of these powerful vision transformers and discover the ability to scale them up into an more powerful model with a low cost.
We demonstrate that RMoE is not restricted to particular vision transformers.

\subsection{Mixture of Experts}
Mixture of Experts (MoE) has a long history through these decades~\cite{chen1999improved,jacobs1991adaptive,jordan1994hierarchical,yuksel2012twenty}.
Various of expert architectures are proposed~\cite{collobert2001parallel,deisenroth2015distributed,shahbaba2009nonlinear,theis2015generative,tresp2000mixtures}.
In the meanwhile, instead of designing experts in the architecture, there are similar research ideas to generate a set of outputs and pick the correct one, such as multi-choice learning~\cite{guzman2012multiple,lee2017confident,lee2016stochastic}.

Recently, Scaling up transformers using MoE is proven effective to achieve the state-of-the-art performance on various of tasks \cite{raffel2019exploring,riquelme2021scaling}. 
Yet, it is still expensive to train a neural network with billions of parameters.
Compared with non-MoE models, an MoE neural network contains a set of experts, e.g., multilayer perceptrons (MLPs), and a router to select which subset of experts are used for each input data point.
It increases the network capacity by such conditional computation while maintaining a relative efficient training.
MoE has been widely used in Natural Language Processing \cite{fedus2021switch,hansen1999combining,lepikhin2020gshard,shazeer2017outrageously}
and computer vision~\cite{abbas2020biased,eigen2013learning,gross2017hard,riquelme2021scaling,wang2020deep,yang2019condconv}.
In our work, we also strive to scale up the model capacity of vision transformers.
Compared with previous works on MoE vision transformers, we develop a more efficient training pipeline, which reduces the total training cost to obtain a large MoE vision transformer.
Such efficiency is achieved via our factorization of the experts' weights.
In addition, we are among the first to solve the problem of applying MoE on downstream tasks with high-resolution images such as segmentation and detection.
These kinds of tasks pose technical challenges on the device memory while requiring a strong backbone for a good performance, which naturally fit our work.

\section{Background}
\label{sec:background}
In this section, we introduce the basic idea of MoE and how to employ it to transformers to scale up the model. 

\noindent \textbf{MoE layer.}
MoE layers are critical components in an MoE model.
By allowing input-dependent conditional computation, different experts in an MoE layer are assigned to process with different parts of the input space~\cite{jacobs1991adaptive}.
As one of the most common setting~\cite{shazeer2017outrageously}, an MoE layer contains two components: (1) $n$ experts $E_i(x) : \mathbb{R}^{D_{\text{in}}} \rightarrow \mathbb{R}^{D_{\text{out}}}$, $i = 1...n$ to process different inputs and (2) a gate function $G(x) : \mathbb{R}^{D_{\text{in}}} \rightarrow \mathbb{R}^{n}$ to route different inputs to different experts.
Given an input $x\in\mathbb{R}^{D_{in}}$, an $n$-experts MoE layer computes the conditional output $y \in \mathbb{R}^{D_{out}}$ as the weighted sum of gate function $G(x)$ and experts outputs $[E_i(x)]$:
\begin{equation}
\label{eq:moe_background}
    y = \sum_{i=1}^{n}G(x)_i E_i(x) .
\end{equation}
$G$ and $E_i$ are usually modeled with neural networks.  
In our research, we follow the previous designs~\cite{riquelme2021scaling,shazeer2017outrageously} to set $G$ as a linear gate with a $softmax$ function. To encourage sparsity in an MoE layer, usually it restricts only $k$ experts where $k < n$ to participate in the computation for each input. So a TopK operator is utilized in the gate function to force only $k$ experts used while others are skipped for an input. 
Thus we can write $G(x)$ as
\begin{equation}
G(x)= \text{TopK}(\textit{softmax}(\theta_g x)),
\end{equation}
where $\theta_g$ is the parameters of the linear operation in a gate.

\noindent \textbf{Load Balancing Loss.}
In practice, one major problem to train an MoE model is that some of the experts are routed with much less data points than other experts.
This will induce insufficient training for the experts routed with less data points.
In addition, as experts are usually processed in parallel, it will also hurt the training efficiency due to the buckets effect.
To avoid this problem, we usually add a load balancing loss~\cite{shazeer2017outrageously} on the gate function to the total training loss.
Given a batch of inputs $X$, a widely-used load balancing loss $L$ is defined as
\begin{equation}
    L(X) = \left( \frac{\text{std}\left(\text{Imp}(X)\right)}{\text{mean}\left(\text{Imp}(X)\right)} \right)^2,
\end{equation}
where $\text{Imp}(X) := \sum_{x\in X}G(x)$.
In practice, we may add this load balancing loss to total training loss with a balance weight $w_{\text{balance}}$.

\noindent \textbf{MoE Transformer.}
To employ MoE to transformers to build MoE transformers, a widely-used approach is replacing some MLP layers in a non-MoE transformer by MoE layers~\cite{riquelme2021scaling}.
In particular, in an MoE layer, the experts share the same structure with the original MLP.
The gate function receives the output from the previous attention layer and routes the representations for tokens to different experts.
The sparse outputs of experts are combined via Eq.~\ref{eq:moe_background}.

\section{Residual Mixture of Experts}

\subsection{Motivation}
\label{sec:motivation}

\begin{figure}[ht]
    \centering
    \includegraphics[width=0.9\textwidth]{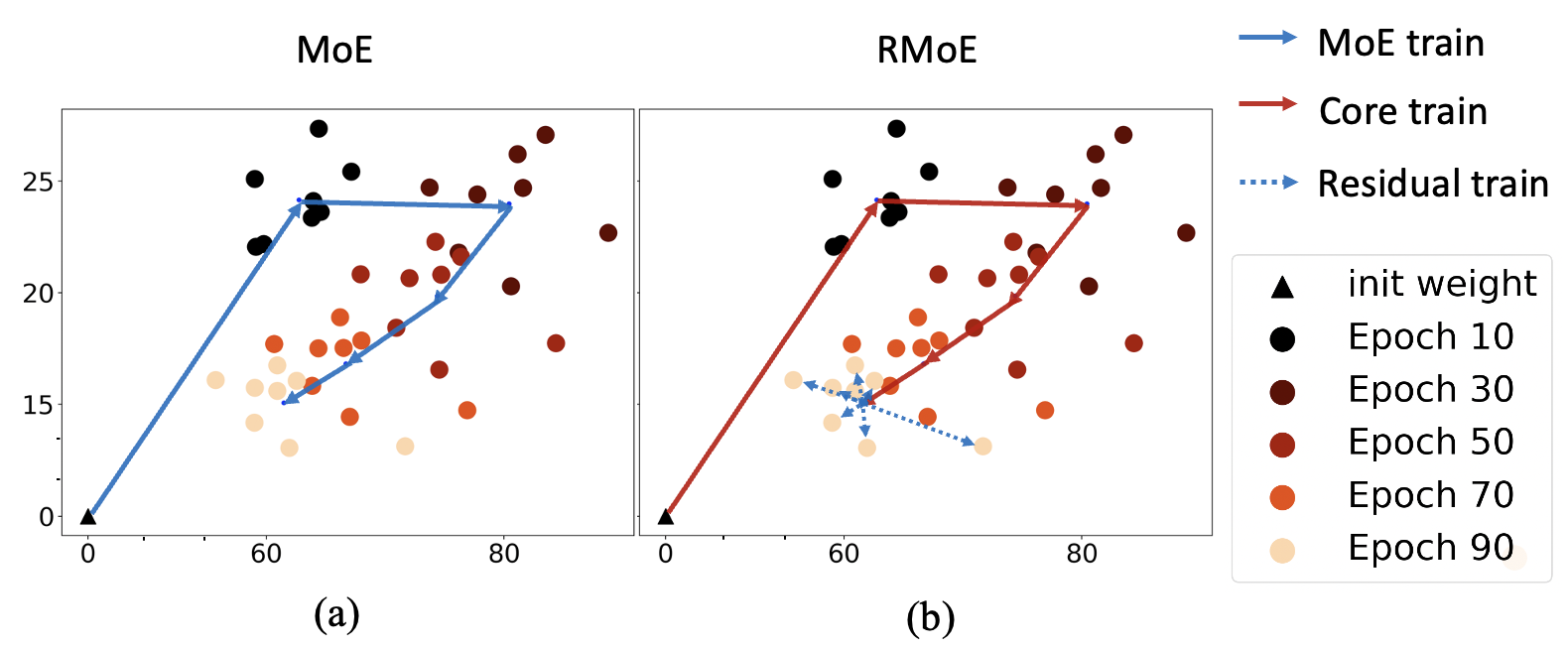}
    \vspace{-1em}
    \caption{Visualization of experts weights evolution during an MoE training process. Here we show the visualization of the experts in the last layer of an 8-experts MoE Swin-T.
    We project the weights to a 2D plane using principle components analysis.
    Each point is the weight of an expert at a training epoch.
    The X and Y axis are the first and second principle components, respectively:
    (a) in the training process, experts weights at different epochs are clustered and the variance of each weight cluster (points of the same color) is much smaller than the variance of cluster centers (blue solid line) during the training.
    Motivated by this pattern, our RMoE on (b) first learns the centers of experts' weights (red solid line) and then learns their residual weights (dash blue line).
    }
    \label{fig:motivation}
    \vspace{-1em}
\end{figure}

Our research is motivated by analyzing how the weights of experts in an MoE vision transformer evolve during the training process.
Specifically, we trained an 8-experts MoE Swin-T~\cite{liu2021swin} on the  ImageNet22K~\cite{deng2009imagenet} for 90 epochs, where we applied MoE to every other Swin transformer block following previous research~\cite{riquelme2021scaling}.
Given the trained model, we focused to analyze the weights of experts in the last MoE layer because: (1) deeper routing decisions correlate with image classes and present richest semantic information~\cite{riquelme2021scaling}; and (2) the last layers affect the most to the classification performance.
The weights of experts at different epochs are visualized by projecting to a 2D plane.
To avoid non-linear distortion in the projection process, we adopted principal components analysis and visualized the top 2 principal components as values on X and Y axis in a scatter-plot.
As shown in Figure~\ref{fig:motivation} (a), we observed that the weights of different experts are clustered according to their training epochs.
More importantly, the variance of each weight cluster is much smaller than the variance of cluster centers during the training.

These observations trigger us to design a more efficient training pipeline than the current MoE training.
In particular, as shown in Figure~\ref{fig:motivation} (b), we can factor the MoE training into two stages.
First we train the centers of expert clusters (solid red path).
Second, we train the residual weights of experts (dash blue paths).
Comparing these two stages, the weights change in the second stage are much smaller than that in the first stage.
Thus, the training cost needed for the second stage is much smaller and the total training cost is mainly determined by the first stage.
In the first stage, to train the weight cluster centers, we can use a non-MoE transformer instead of an MoE model because the experts center is input independent.
Thus, the first training stage itself is efficient because the training cost for an MoE vision transformer is $1.5 \sim 2\times$ higher than that of non-MoE models~\cite{riquelme2021scaling}.
In practice, for downstream tasks, we can often even skip the first training stage and only perform the residual training by directly using the available non-MoE vision transformer checkpoints.
As a result, we can train a large MoE vision transformer efficiently.

\subsection{Formulation}

Generally, the conditional computation of an MoE model $f$ can be formulated as $y = f(x; \theta(x))$, $\theta(x) \subset \Theta$, where $x$ is an input data point, $y$ is the output and $\Theta$ is the whole set of model parameters.
Using this formulation, the training of an MoE model aims at minimizing empirical loss $\mathcal{L}$:
\begin{equation}
    \mathop{\text{min}}\limits_{\theta} \sum_{x \in X} \mathcal{L}(f(x; \theta(x))).
\end{equation}
where $X$ is the training set.

Motivated by the observations in Sec.~\ref{sec:motivation}, we factor the conditional computation $\theta(x)$ as an input-independent  $\bm{\theta_0}$ and an input-dependent residual $\theta_r(x)$:
\begin{equation}
    \theta(x) = \bm{\theta_0} + \theta_r(x).
\end{equation}
In this factorization, the size of the weights residual $\theta_r(x)$ is \textbf{much larger} than that of the weights core $\bm{\theta_0}$ because experts in a MoE layer share the same weights core and we often have $8 - 32$ experts in an MoE vision transformer~\cite{riquelme2021scaling}. 
On the contrary, the weights residual can be efficiently trained with \textbf{much less} computation resource, e.g., less training data and less training epochs.

Using this factorization, we propose the Residual Mixture of Experts (RMoE) training pipeline, which formulates the training of an MoE vision transformer as a bilevel optimization with residual $\theta_r$ as the upper-level variable and core $\bm{\theta_0}$ as the lower-level variable:
\begin{alignat*}{2}
& \mathop{\text{min}} \limits_{\theta_r}  && \sum \mathcal{L}(f(x; \bm{\theta_0}^* + \theta_r(x)))\\
& \text{s.t.} && \bm{\theta_0}^* = \mathop{\text{argmin}}\limits_{\bm{\theta}} \sum  \mathcal{L}(f(x; \bm{\theta}))\}.
\end{alignat*}

\subsection{Our Designs}

To fully leverage the benefits of the RMoE training pipeline, there are several practical designs considerations needed to discuss:

\noindent \textbf{Training Pipeline Design.} The training pipelines of non-MoE and MoE models are similar and can be divided into several stages.
As shown in Figure~\ref{fig:intro} (a) and (d), in the first stage, vision transformers are pretrained on a large-scale upstream dataset.

Then, there is an optional intermediate finetune stage.
It is to bridge the task gap or resolution gap between upstream and downstream, e.g., BeiT~\cite{bao2021beit}.

Finally, the model is finetuned on the downstream dataset, such as semantic segmentation or object detection.

Compared with non-MoE and MoE training pipelines, RMoE starts with the non-MoE transformer training on the upstream task and efficiently finetunes the non-MoE transformer into a MoE transformer.
As shown in Figure~\ref{fig:intro} (b) and (c), in RMoE training, we can either intermediately finetune the MoE model on the upstream dataset for a few epochs or directly finetune the MoE model on the downstream dataset.

We denote these two different pipelines as RMoE-I (Intermediate) and RMoE-D (Downstream).

\noindent \textbf{Performance-preserving between non-MoE and MoE transformers.}
In RMoE, we need to inherit the learned weight core from the upstream non-MoE pretraining to perform a residual learning.

To this end, we initialize the weights of each expert in an MoE layer as the weights learned in the corresponding MLP layer from non-MoE pretraining.
In addition, we add a small noise to each expert to escape from the local minima.

In practice, we find that that directly inheriting the experts weights will cause performance degradation.
Specifically, as in Eq.~\ref{eq:moe_background}, the output of an MoE layer is the weighted sum of gate function $G(x)$ and experts outputs $[E_i(x)]$:
$MoE(x) = \sum_{i=1}^{n}G(x)_i E_i(x).$
In RMoE, because all experts inherit the weights from non-MoE pretraining and $G(x)$ is post-processed by a Top-K operation to ensure sparsity, we have:
\begin{equation}
    \text{MoE}(x) \approx (\sum_{i=1}^{n}G(x)_i ) \cdot \text{MLP}(x) < \text{MLP}(x).
\end{equation}
Thus, the outputs of MoE layers are scaled down compared with original MLP layers and cause performance degradation.

To tackle this issue, we propose to align the outputs of the MoE and corresponding MLP layers while maintain the gradient flow in an MoE layer:

\begin{equation}
    \label{eq:stop}
    y =  \sum_{i=1}^n \text{StopGrad}((1 - G(x)_i) E_i(x)) +  G(x)_i E_i(x),
\end{equation}
where $\text{StopGrad}$ operation is to stop the gradient of the given term.
In this way, when back-propagate the gradient, the experts can still get a normal gradient update only for the top-K experts. In RMoE, we only apply this for intermediate finetune because: (1) the intermediate finetune stage in RMoE is often short with a small learning rate and it is hard to fully recover the performance degradation; and (2) the downstream finetuning with the new decode head is done with a large learning rate and the performance drop will recover quickly.

\noindent \textbf{Layer Choices for MoE layers.} As introduced in Sec.~\ref{sec:background}, MoE layers are the key difference between MoE and non-MoE transformers.
Thus, to finetune a pretrained non-MoE transformer into an MoE transformer, we need to replace several MLP layers to MoE layers~\cite{fedus2021switch,riquelme2021scaling}.
To avoid potential over-fitting, we only select the most important layers that contribute most to network training.

Inspired by Firefly Splitting~\cite{wu2021firefly}, in RMoE, we select the layer that can maximally decrease the loss function in a layer-wise fashion.
First, we over-grow the non-MoE transformer by replacing all MLP layers with MoE layers.
After the over-growing, we calculate the loss decrease.
In particular, given $f$ representing a vision transformer with $L$-layers.
Thus, $\mathcal{L}_{t}(f_{\text{non-MoE}})$ is the training loss for the pretrained non-MoE vision transformer and $\mathcal{L}_{t}(f_{\text{RMoE}}; \bm{\theta_r})$ represents the loss after introducing MoE layers into a non-MoE model in RMoE training, with residual weight $\bm{\theta_r}$.

As we mentioned before, we add a small noise over weight of MLP to initialize the experts weights, it is equivalent to initialize $\bm{\theta_r} \leftarrow \epsilon\bm{\theta_0}$, where $\epsilon$ is a small enough value that only perturbs the network and output in a small range.
Thus, the loss decrease can be decomposed via Taylor approximation as:
\begin{equation}
\label{eq:firefly}
     \mathcal{L}_{t}(f_{\text{RMoE}};\bm{\theta_r}) = \mathcal{L}_{t}(f_{\text{non-MoE}}) + \epsilon \sum_{l=1}^{L}  \theta_r^ls_l + \epsilon ^2 O(\bm{\theta_r}).
\end{equation}
\begin{equation*}
     s_l \approx \nabla_{\theta_r^l} \mathcal{L}_{t}(f_{\text{RMoE}};\bm{\theta_r} ) .
\end{equation*}
Next, we find the most important $N$ layers to decrease the loss.
We achieve this by first optimizing the initial of $\bm{\theta_r}$ for several gradient descent step and because the residual weight $\bm{\theta_r}$ is initialized with a small enough factor $\epsilon$:
\begin{equation}
\label{eq:solution}
    \bm{\hat{\theta}_r} = \mathop{\text{argmin}}\limits_{\bm{\theta_r}} \{ \epsilon\sum_{l=1}^{L}  \theta_r^ls_l \quad \text{s.t.} \quad ||\bm{\theta_r}||_0 \le N \},
\end{equation}
where  $||\bm{\theta_r}||_0 := \sum_{l=1}^{L}  \mathbb{I} (\theta_r^l \neq 0) $.
To select the best layers to maximally decrease loss function, the optimal solution is  selecting the layer with largest $N$ gradient magnitude $|s_l|$. In practice, we want to get a general scaled up model at once for all the downstream tasks, so we calculate the the gradient magnitude on all the target downstream tasks and select the $N$ layers with top high scores sum among all the tasks.

\section{Experiment}
\label{sec:exp}
In this section, we evaluate the effectiveness and efficiency of RMoE.

We first deliver a comprehensive comparison of different training pipelines, i.e., non-MoE, RMoE, and MoE.
Then, to show the capability of RMoE to train large vision transformers, we use RMoE to train models using Swin-L and BeiT-L~\cite{bao2021beit} backbones.
Finally, we conduct a set of ablation studies to compare different design choices of RMoE.

\begin{table*}[ht]
    \centering
     \setlength{\tabcolsep}{0.9mm}
    \renewcommand\arraystretch{1.4}
    \scalebox{0.95}{
    \begin{tabular}{l|c|c c c c |c c c}
    \Xhline{3\arrayrulewidth}
        \multirow{2}{*}{Model}      &  \multirow{2}{*}{Type}     &\multicolumn{3}{c}{ADE20K} & GPU Days & \multicolumn{2}{c}{MS-COCO} & GPU Days \\
         &  &   mIoU  & mIoU (ms+flip) &  Params & Scratch / Pretrained & AP  & Params & Scratch / Pretrained \\
    \hline
        \multirow{5}{*}{Swin-T} & Non-MoE   &  44.6 & 45.9  & 60M & 56.3 / 6.3 & 41.6  & 39M & 52.6 / 2.6\\

                \cline{2-9}
                & \cellcolor{Graylight}RMoE-D &  \cellcolor{Graylight} 45.3 & \cellcolor{Graylight}46.6  & \cellcolor{Graylight}103M & \cellcolor{Graylight}56.8 / 6.7 & \cellcolor{Graylight}42.6 & \cellcolor{Graylight}82M & \cellcolor{Graylight}53.0 / 2.9 \\
              & \cellcolor{Graylight}RMoE-I (1k)  &  \cellcolor{Graylight}45.6 & \cellcolor{Graylight}46.9  & \cellcolor{Graylight}103M & \cellcolor{Graylight}57.3 / 7.1  & \cellcolor{Graylight}42.9 & \cellcolor{Graylight}82M  & \cellcolor{Graylight}53.5 / 3.4 \\
         & \cellcolor{Graylight}RMoE-I  &  \cellcolor{Graylight}45.7 & \cellcolor{Graylight}47.2  & \cellcolor{Graylight}103M & \cellcolor{Graylight}57.7 / 7.5  & \cellcolor{Graylight}43.0 & \cellcolor{Graylight}82M  & \cellcolor{Graylight}53.9 / 3.8 \\
          \cline{2-9}
          & MoE  &  45.9 & 47.3 & 120M & 75.1 / 75.1 & 43.1 &  99M & 71.8 / 71.8  \\

                          \Xhline{3\arrayrulewidth}
        \multirow{5}{*}{CvT-13} & Non-MoE   &  44.9 & 46.4  &  45M  & 46.2 / 6.2  & 38.3 & 29M & 42.4 / 2.4 \\

                     \cline{2-9}
                     & \cellcolor{Graylight}RMoE-D  &  \cellcolor{Graylight}45.4 & \cellcolor{Graylight}46.9    & \cellcolor{Graylight}94M & \cellcolor{Graylight}46.8 / 6.6 & \cellcolor{Graylight}39.6  &  \cellcolor{Graylight}78M & \cellcolor{Graylight}42.9 / 2.6 \\
                     & \cellcolor{Graylight}RMoE-I (1k)  &  \cellcolor{Graylight}45.6 & \cellcolor{Graylight}47.0 &  \cellcolor{Graylight}94M & \cellcolor{Graylight}47.3 / 7.0 & \cellcolor{Graylight}39.7 &  \cellcolor{Graylight}78M & \cellcolor{Graylight}43.5 / 3.3  \\
                    & \cellcolor{Graylight}RMoE-I  &  \cellcolor{Graylight}45.8 & \cellcolor{Graylight}47.2 & \cellcolor{Graylight} 94M & \cellcolor{Graylight}47.6 / 7.3 & \cellcolor{Graylight}39.9 &  \cellcolor{Graylight}78M & \cellcolor{Graylight}43.8 / 3.6  \\
                     \cline{2-9}
                       & MoE &  46.0 & 47.4   &  119M & 60.4 / 60.4 & 40.1 &  103M & 56.1 / 56.1 \\

    \Xhline{3\arrayrulewidth}

    \end{tabular}
    }
    \vspace{0.1em}
    \caption{A comprehensive comparison between different training pipelines including non-MoE, MoE, RMoE-I and RMoE-D on ADE20K segmentation task and MS-COCO object detection task. RMoE-I (1k) represents intermediate finetune the model on ImageNet 1k instead of ImageNet 22k. Non-MoE represents using the original transformers as backbones. The GPU Days measures the total training time in one Nvidia Tesla V100-32GB GPU. `Scratch' represents that the pipeline is fully performed from upstream pretrain to downstream finetune.
    `Pretrained' means loading from existing pretrained upstream \textbf{non-MoE} checkpoints, which is usually available in the computer vision community. Notice that Swin-T and CvT-13 has a different batch size during upstream training, so the GPU Days does not measure the speed relationship bewteen two different backbones.}
    \vspace{-1em}
    \label{tab:main}
\end{table*}

\subsection{Comparing Different Training Pipelines}
\label{exp1}
We first compare different training pipelines in Figure~\ref{fig:intro}, i.e., non-MoE, MoE, RMoE-I and RMoE-D to demonstrate the effectiveness of RMoE.
In each training pipeline, we used the image classification on ImageNet22k~\cite{deng2009imagenet} as the upstream pretraining.
We adopted two representative downstream tasks, i.e., semantic segmentation and object detection.
The model performance on semantic segmentation and object detection are evaluated on ADE20K~\cite{zhou2019semantic} and MS-COCO~\cite{lin2014microsoftcoco}, respectively.
For backbones, we selected two representative vision transformers, Swin-T and CvT-13.
For decoder head, we employ UperNet~\cite{xiao2018upernet} and RetinaNet~\cite{lin2017focal} for segmentation and detection tasks.

\noindent \textbf{Training Setting.} For upstream pretraining, we followed the training strategy in the original papers~\cite{liu2021swin,wu2021cvt}.
Specifically, all upstream tasks were trained for 90 epochs with the input resolution $224 \times 224$. We set the batch size to 1024 for Swin-T and 2048 for CvT-13 following the original settings and optimized them using AdamW~\cite{loshchilov2019adamw} optimizer with an initial learning rate $0.001$ for Swin-T and $0.01$ for CvT-13. A cosine learning decay scheduler is used along with the training. The weight decay is $0.05$.

For intermediate finetuning stage of RMoE-I, we reduced the non-MoE model training epochs for 5 and finetuned the MoE model on ImageNet22K for additional 5 epochs to align the total number of epochs with other settings. The initial learning rate was set to $0.0001$ and a cosine learning rate decay was used. Other setting are the same as the upstream training.
\lm{We also apply the intermediate finetune on ImageNet 1k for 30 epochs and $384 \times 384$ resolution for a more viable setup. We mark it as RMoE-I (1k) in Table \ref{tab:main}.}
The settings in downstream finetuning stage are the same with those in Swin Transformer for both Swin-T and CvT-13 on the semantic segmentation task expect a longer training steps 160k $\rightarrow$ 200k for a fully converge for RMoE.
We applied the same training strategy for all the models on ADE20K.
For object detection on MS-COCO, we trained all the models using AdamW optimizer with learning rate $0.0001$ and regular $1\times$ schedule without multi-scaled augmentation.

\noindent \textbf{MoE Settings.} For all the MoE models, we used 8 MLP experts with a switch gate~\cite{fedus2021switch}, which means that each token is only routed to one expert in an MoE layer.
For MoE training, we followed the settings of the state-of-the-art MoE training pipeline ~\cite{riquelme2021scaling}.

Specially, for CvT-13, we add MoE layers after stage 2 to stabilize the training.

For RMoE training, we add a random noise $\epsilon = 0.01$ on each expert after initialized from a non-MoE model to encourage diversity.

We calculated the gradient-based grow score and selected top-3 score layers to apply MoE.
The detailed layer selections can be found in Table~\ref{tab:expand}.
For both MoE and RMoE training, we applied the load balance loss in Sec.~\ref{sec:background} with a weight $w_{\text{balance}} = 0.01$ during upstream training and intermediate finetune, a $w_{\text{balance}} = 0.0001$ weight is added during downstream finetune.

\noindent \textbf{Experiment Results.} As we can see in Table~\ref{tab:main}, the original MoE training gets the best performance on all of the tasks and models but with highest training cost.
The performance of RMoE-I is comparable to MoE with saving around 30\% training cost (training from scratch).
RMoE-D performs slightly worse than RMoE-I, but both outperform the non-MoE training with minor training cost increase.
On Swin-T and CvT-13 backbones, compared with non-MoE baselines, RMoE-I gets +1.1 / 0.9 mIoU.
It also gets + 1.4 / 1.6 AP on the detection task.
RMoE-D, which directly uses experts on the downstream tasks, can still gets +0.7 / 0.5 mIoU and +1.0 / 1.3 AP.
In practice, we can leverage the shared checkpoints of Swin-T and CVT-13 to further reduce the training cost.
In that case, we can use RMoE to scale up a vision transformer for downstream tasks with less than $10\%$ of the training cost of the MoE training.

\begin{table}[ht]
    \centering
    \setlength{\tabcolsep}{1.0mm}
    \renewcommand\arraystretch{1.2}
    \begin{tabular}{l|c|c|c|c}
    \Xhline{3\arrayrulewidth}
        Stage & Swin-T & CvT-13 & Swin-L & BeiT-L   \\
   \Xhline{3\arrayrulewidth}
         Stage 1  &  - & - &  2  &  \makecell{8, 10, 12, 16,\\ 17, 18, 20, 24} \\
         \hline
         Stage 2 & 2 & 2 & 2 & N/A \\
         \hline
         Stage 3 & 6 &  3,  9 &  9, 12, 18 & N/A \\
         \hline
         Stage 4 & 2 & N/A & 2 & N/A \\
     \Xhline{3\arrayrulewidth}
    \end{tabular}
    \vspace{0.1em}
    \caption{Layer selections in RMoE training for downstream tasks. For example, in Swin-T, we select the 2nd layer in stage 2,4 and the 6th layer in stage 3 as MoE layers.
    Notice that CvT only has 3 stages, and BeiT does not have stages.
    So for alignment, we represent CvT13 using Stage 1,2,3 and BeiT-L only using Stage 1. }
    \vspace{-1.5em}
    \label{tab:expand}

\end{table}

\begin{figure}
\centering
    \centering
    \includegraphics[width=0.75\textwidth]{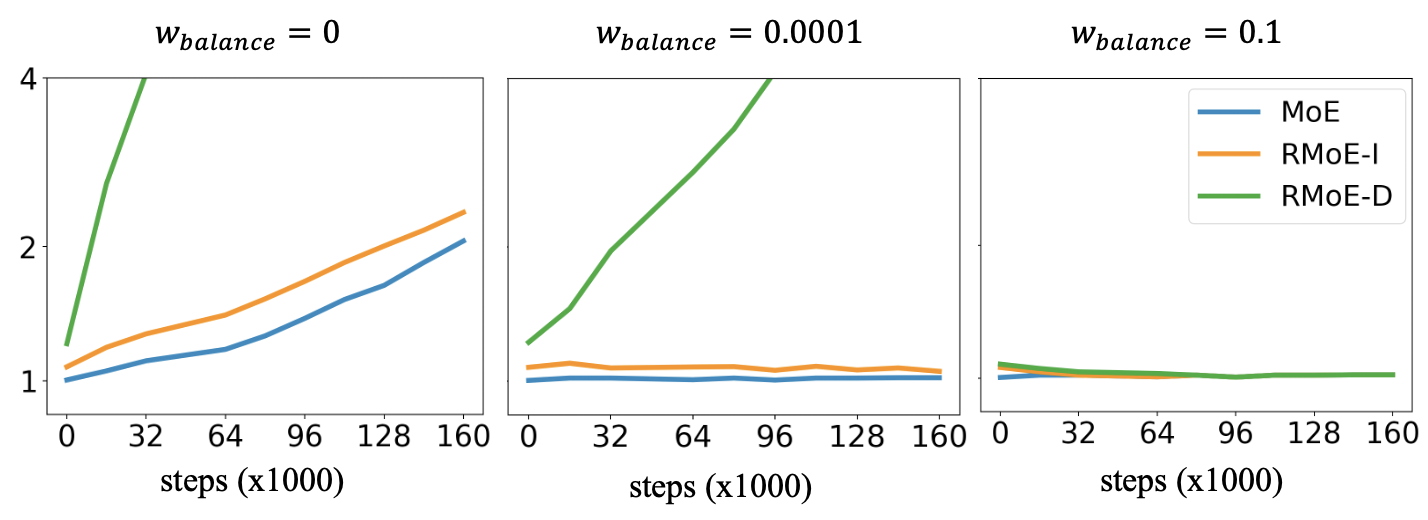}

    \vspace{-1em}
    \caption{Balance loss on the ADE20K segmentation tasks with different loss weights. Weight=0 denotes removing the balance loss. Both RMoE-I and MoE can get a good balance under a small weight, but RMoE-D can only get balanced when a large balance loss weight is used.}

    \label{fig:bal}
\end{figure}

\noindent \textbf{Analysis of the Gap between RMoE-I and RMoE-D.}
The balancing loss is found to be critical for explaining the performance difference between RMoE-I and RMoE-D.
In the finetuning stage of MoE training pipeline, load balancing loss is often not added.
For example, in V-MoE, it is claimed that the well-trained router can balance each expert well without a balancing loss.
However, in RMoE-D, we find that the newly introduced decoder head and a large initial learning rate in downstream tasks can break the load balance among experts.
To demonstrate this, we compare the training processes of MoE, RMoE-D and RMoE-I w/ and w/o the balance loss in the downstream finetuning stage.
As shown in Figure~\ref{fig:bal}, we find that w/o balance loss, all the methods, including MoE, result in a bad load balance. When we add the balance loss with a small weight $w_{\text{balance}} = 0.0001$, the load balance in the RMoE-D is still bad.
RMoE-I, which benefits from an intermediate finetune, can smoothly decrease the balance loss.
A further large weight $w_{\text{balance}} = 0.1$ can make RMoE-D balanced but it would hurt the performance as well.

\begin{figure}[t]
 \centering
    \includegraphics[width=0.70\textwidth]{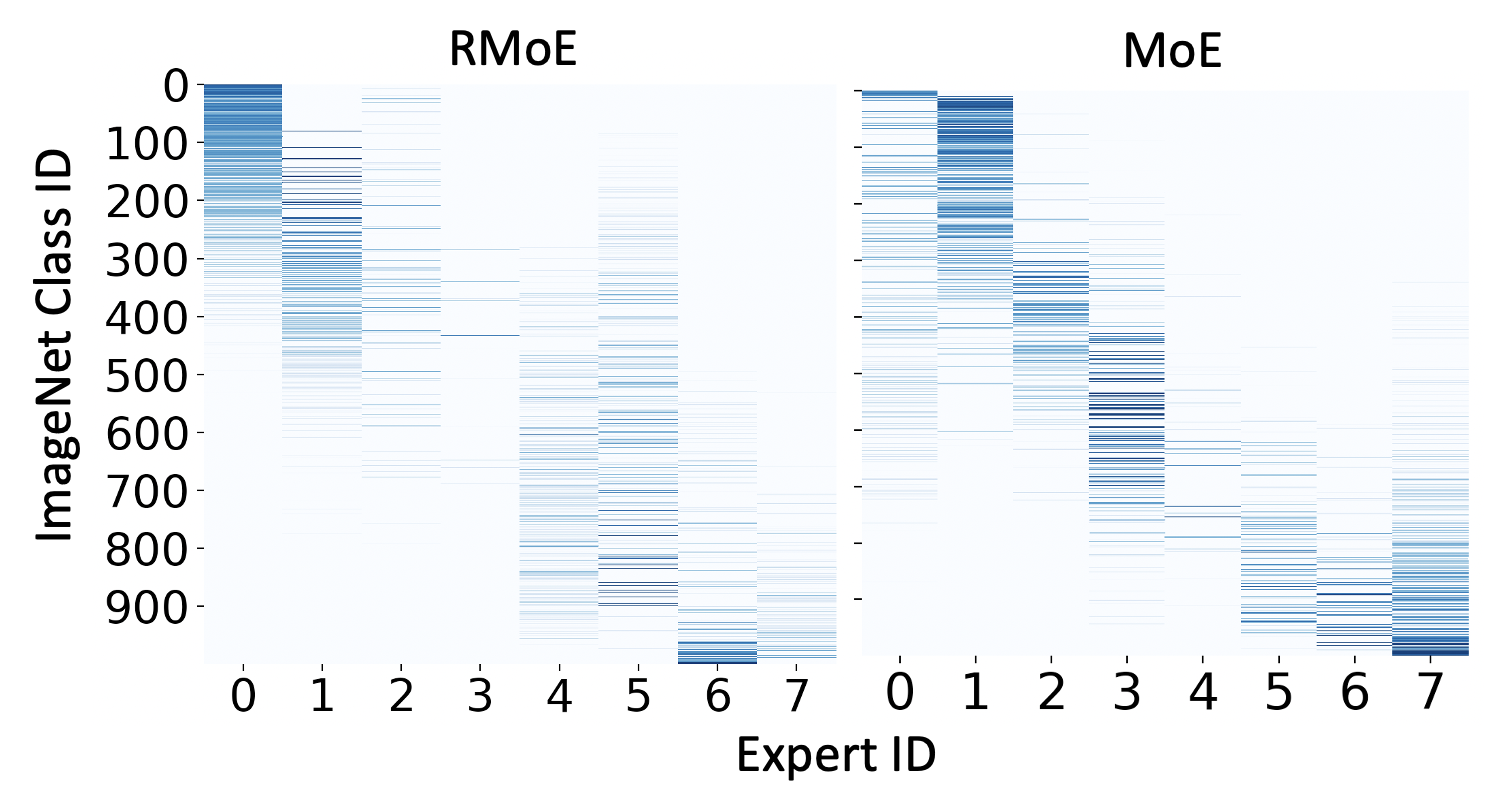}
    \vspace{-1em}
    \caption{Visualization of experts specialization of RMoE and MoE training. A darker color in a cell indicates that the expert is more specialized to tackle the images in that class of ImageNet. We separately reorder the ImageNet classes for RMoE and MoE to get a better visualization. }
    \label{fig:special}
    \vspace{-1em}
\end{figure}

\noindent \textbf{Expert Specialization.}
To analyze how experts distribute different images, we visualize how many images of a given ImageNet class use each expert in Figure~\ref{fig:special}.
In particular, each row is an ImageNet class and each column is an expert.
The color in one cell denotes the average routing weight (the largest $G(x)$ in Eq.~\ref{eq:moe_background}) for all the tokens in a specific class for an expert.
Thus, darker color in a cell means that the expert is specialized to tackle the images in that class.
We compared both RMoE and MoE training using RMoE-I (1k) intermediate finetune setting and selected the last MoE layer to analyze.
We find that experts in both RMoE and MoE training are specialized across the classes in ImageNet.
It echos our observation in Sec.~\ref{sec:motivation} that, although RMoE uses much less computation resource to train the experts, expert specialization is achieved by both RMoE and MoE training.




\subsection{Using RMoE to Train Large Transformers}

We further demonstrate the capability of RMoE to further scale up large vision transformers, including Swin-L and BeiT-L, because they achieve state-of-the-art results on downstream tasks.

In particular, we applied RMoE to Swin-L w/ UperNet \cite{xiao2018upernet} and Maskformer \cite{cheng2021maskformer}, BeiT-L w/ UperNet for ADE20K segmentation and Swin-L w/ HTC++~\cite{chen2019hybrid,liu2021swin} on MS-COCO object detection.

\noindent \textbf{Experiment Setup.}  Training hyper parameters are the same with those in the finetuning setting of the Swin-L, Maskformer and BeiT-L.
For the HTC++ decoder head, we reduced the schedule from $6\times$ to $3\times$ since the scaled-up model has a better converging speed.
For intermediate finetune, the weights of experts directly inherit from the official pretrained checkpoint \footnote{https://github.com/microsoft/Swin-Transformer}
\footnote{https://github.com/microsoft/unilm/tree/master/beit}. The rest of the settings keep the same as in Section~\ref{exp1}.
For Swin-L, we chose 6 layers and for BeiT-L we chose 8 layers as MoE layers  using the gradient-based grow score.
The detailed layer selections are shown in Table~\ref{tab:expand}.
Other settings related to RMoE are the same as Section~\ref{exp1}.

\noindent \textbf{Experiment Results.}
As we can see from Table~\ref{tab:seg}, RMoE-I can further scale up the backbones to larger model capacity, with a performance gain at around 1.0 on Swin-L segmentation,
We also improve BeiT-L segmentation result by 0.4.
RMoE-D can also get a +0.7 improvement compared with the non-MoE model for a nearly free cost on Swin-L model with different Head.
For the object detection task, the results shown in Table~\ref{tab:od} indicate that RMoE-I and RMoE-D get a +0.6 / 0.4 higher AP compared with non-MoE model baseline with only half finetuning epochs.

\begin{table}[t]
    \centering
        \setlength{\tabcolsep}{2.0mm}
    \renewcommand\arraystretch{1.3}
    \begin{tabular}{l|c|c|c|c c}
 \Xhline{3\arrayrulewidth}
      Backbone & Head  & Type & Params &  mIoU & \makecell{mIoU \\ (ms+flip)} \\
  \hline
       \multirow{3}{*}{Swin-L}  &  \multirow{3}{*}{UperNet}  & Non-MoE & 234M & 52.0 & 53.5 \\
        &  & \cellcolor{Graylight}RMoE-D & \cellcolor{Graylight}476M & \cellcolor{Graylight}52.7  & \cellcolor{Graylight}54.2 \\
       & & \cellcolor{Graylight}RMoE-I & \cellcolor{Graylight}476M & \cellcolor{Graylight}\textbf{52.9} & \cellcolor{Graylight}\textbf{54.4} \\

   \hline
       \multirow{3}{*}{Swin-L}  & \multirow{3}{*}{\makecell{Mask\\Former}}  & Non-MoE & 212M & 54.0$^{\dag}$ &  55.6 \\
          & & \cellcolor{Graylight}RMoE-D & \cellcolor{Graylight}454M & \cellcolor{Graylight}54.7 & \cellcolor{Graylight}56.3 \\
     & & \cellcolor{Graylight}RMoE-I & \cellcolor{Graylight}454M & \cellcolor{Graylight}\textbf{55.0} & \cellcolor{Graylight}\textbf{55.7} \\

   \hline
       \multirow{3}{*}{BeiT-L}   &  \multirow{3}{*}{UperNet} & Non-MoE & 502M  & 56.4$^{\dag}$ & 57.0 \\
         &  & \cellcolor{Graylight}RMoE-D & \cellcolor{Graylight}737M & \cellcolor{Graylight}56.7  & \cellcolor{Graylight}57.1 \\
    &   & \cellcolor{Graylight}RMoE-I & \cellcolor{Graylight}737M & \cellcolor{Graylight}\textbf{56.9} & \cellcolor{Graylight}\textbf{57.4} \\

 \Xhline{3\arrayrulewidth}

    \end{tabular}
       \vspace{0.1em}
    \caption{ADE-20K result on various of large backbones and decoder heads. $^{\dag}$ Our result on MaskFormer and BeiT-L UperNet without TTA is slightly lower than the result reported in original papers. However, the TTA results keep the same.}
    \label{tab:seg}
    \vspace{-1.0em}
\end{table}

\begin{table}[t]
    \centering
        \setlength{\tabcolsep}{2.0mm}
    \renewcommand\arraystretch{1.2}
    \begin{tabular}{l|c|c|c c}
    \Xhline{3\arrayrulewidth}
       Type & Head & Params  &  AP$^{\text{box}}$ & AP$^{\text{mask}}$ \\
  \Xhline{3\arrayrulewidth}
       Non-MoE  &  HTC++ (6x) & 284M  & 57.1 & 49.5 \\
        \cellcolor{Graylight}RMoE-D  & \cellcolor{Graylight}HTC++ (3x) & \cellcolor{Graylight}526M   & \cellcolor{Graylight}57.5 & \cellcolor{Graylight}49.8 \\
       \cellcolor{Graylight}RMoE-I  & \cellcolor{Graylight}HTC++ (3x) & \cellcolor{Graylight}526M  & \cellcolor{Graylight}\textbf{57.7} & \cellcolor{Graylight}\textbf{50.0} \\

\Xhline{3\arrayrulewidth}

    \end{tabular}
    \vspace{0.1em}
    \caption{Non-MoE and RMoE result of MS-COCO object detection on Swin-L and HTC++. }
    \label{tab:od}
    \vspace{-1.5em}
\end{table}

\subsection{Inference Speed and FLOPs}

\lm{An appealing property for MoE transformer is that it enlarges the model capacity (number of parameter) by only introducing minimal additional inference time and FLOPs than non-MoE transformerts.
RMoE also has this property.
As shown in Table~\ref{tab:inference}, we analyzed the the inference time and FLOPs of Swin-T and Swin-L on ADE20K segmentation tasks using UperNet decoder head.
We show that RMoE has an obvious performance gain but only introduces minimal additional inference time, which is critical in machine learning models deployment.
}
\begin{table}[t]
    \centering
        \setlength{\tabcolsep}{2.0mm}
    \renewcommand\arraystretch{1.3}
    \begin{tabular}{l|c|c|c c}
     \Xhline{3\arrayrulewidth}
        Backbone & Type & mIoU & Inference time (img/s) & FLOPs \\
        \hline
        \multirow{2}{*}{Swin-T} &  Non-MoE  &  44.6   &  29.0 & 945G\\
          &   \cellcolor{Graylight}RMoE-I  &    \cellcolor{Graylight}45.7    &   \cellcolor{Graylight}27.2 & \cellcolor{Graylight}945G    \\
         \hline
          \multirow{2}{*}{Swin-L}   &   Non-MoE    &   52.0     & 16.8 & 3230G \\
                &   \cellcolor{Graylight}RMoE-I &  \cellcolor{Graylight}52.9  &  \cellcolor{Graylight}15.4 &  \cellcolor{Graylight}3230G \\
       \Xhline{3\arrayrulewidth}
    \end{tabular}
    \vspace{0.1em}
    \caption{Comparison of mIoU against inference time and FLOPs between non-MoE, RMoE-I on ADE20K with UperNet decode head. We test the inference speed on a single RTX3090.}
    \vspace{-1em}
    \label{tab:inference}
\end{table}

\subsection{Ablation Study}
\label{sec:ab}
We performed a series of ablation studies for the RMoE to compare different design choices.
The ablation studies were conducted on Swin-T using UperNet on ADE20K segmentation and RetinaNet on MS-COCO object detection.
We used RMoE-I as the baseline due to its better performance and training cost trade-off.

\noindent \textbf{MoE layer selection.}
As one of the key factors, the MoE layer selection in the transformers influences the final performance and model size.
We report different layers choices in Table~\ref{tab:ablayer}. In those settings, Last-2 and Every-2 are introduced in~\cite{riquelme2021scaling}. Last-2 places MoE in the last 2 even blocks and Every-2 places MoE in every other layer. We also introduce the Every-Last, which expands the layer in the last layer of each stage in Swin-T Transformer with the intuition that the layer closer to the decode head is more important. Table~\ref{tab:ablayer} shows the result for different strategies. We find that our gradient score-based selection performs better than Last-2 and Every-2 while keeping the same performance compared with Every-Last. For downstream tasks, every stage's last layer feature will be input into the decode head, so Last-2 is no longer a good choice. It also explains the good performance of Every-Last. Besides, we further optimize Every-last by one layer less, showing the advantages of score-based method. On large backbones, Every-Last can only use four MoE layers, which may be insufficient. For Every-2, the performance slightly drops compared with ours. We think that it may be because Every-2 needs three additional MoE layers while some of the intermediate layers can not contribute a lot to the performance.

\noindent \textbf{Number of Experts and Top-K Gate.}
As common hyper-parameters, the number of experts $n$ and the $k$ of the top-$k$ operation in a gate are usually considered in the MoE experiment. In Table~\ref{tab:kn} we compare our setting with other settings, i.e., $n=4, 16$ and $k=2$.
We see that $k=2$ improves the performance on both the downstream tasks while increasing training time for $10\%$ and a little more FLOPs. With the fewer experts, $n=4$ decreased the performance brought by scaled-up with RMoE-I for around 0.3 on ADE20K and MS-COCO. $n=16$ does not improve the performance, which may differ from the conclusions in other MoE literature.
We speculate that this is mainly due to the load balancing loss used~\cite{riquelme2021scaling}.
It may hurt each expert's performance in the finetuning stage.
A more dedicated finetuning process with improved balancing loss needs to be designed when the number of experts increases, e.g., 128 in \cite{fedus2021switch} or 32 in
\cite{riquelme2021scaling}.

\noindent \textbf{Stop Gradient.}
We apply the stop gradient technique in Eq.~\ref{eq:stop} for RMoE-I intermediate finetune to avoid a performance drop. As shown in Table~\ref{tab:stop}, the performance drops a lot without the stop gradient in the intermediate finetune.

\noindent \lm{\textbf{Noise Initialization Scale.}
To encourage the specialization for each experts after applying RMoE, we apply a certain noise level on each copied weight. Table~\ref{tab:noise} shows how different noise level affects the final result.}

\begin{table}[t]
\begin{minipage}[t]{0.45\linewidth}
    \centering
        \setlength{\tabcolsep}{2.0mm}
    \renewcommand\arraystretch{1.0}
    \scalebox{0.8}{
    \begin{tabular}{l|c|c|c|c}
\Xhline{3\arrayrulewidth}
        & \# Layers & \makecell{Backbone \\ Params} & mIoU & AP  \\
 \hline
        \cellcolor{Graylight}Ours & \cellcolor{Graylight}3 & \cellcolor{Graylight}71M  & \cellcolor{Graylight}\textbf{45.7} & \cellcolor{Graylight}\textbf{43.0} \\
        \hline
        Last-2 & 2 & 69M & 45.3  & 42.4  \\
        Every-2 & 6 & 88M & 45.4 & 42.9  \\
        Every-Last & 4 & 72M & \textbf{45.7} & \textbf{43.0} \\
     \Xhline{3\arrayrulewidth}
    \end{tabular}
    }

    \caption{Ablation study on different MoE layers selections.}
    \vspace{-1em}
    \label{tab:ablayer}
    \end{minipage}
    \hspace{2em}
    \begin{minipage}[t]{0.45\linewidth}
        \centering
        \setlength{\tabcolsep}{2.0mm}
    \renewcommand\arraystretch{1.0}
    \scalebox{0.8}{
    \begin{tabular}{c c|c c|c|c}
\Xhline{3\arrayrulewidth}
     k & n &  \makecell{Backbone \\ Params} & \makecell{Backbone \\ FLOPs} &  mIoU  & AP   \\
 \hline
        \cellcolor{Graylight}1 & \cellcolor{Graylight}8 & \cellcolor{Graylight}71M & \cellcolor{Graylight}4.7G  & \cellcolor{Graylight}45.7  & \cellcolor{Graylight}43.0  \\
        2 & 8 & 71M & 7.4G & \textbf{45.8}  & \textbf{43.3}  \\
        \hline
        1 & 4 & 47M & 4.7G & 45.4  & 42.7\\
        1 & 16 & 121M & 4.7G & 45.7  & 42.9 \\
     \Xhline{3\arrayrulewidth}
    \end{tabular}
    }

    \caption{Ablation study on different top-$k$ gate and number of experts $n$.}
    \vspace{-1em}
    \label{tab:kn}
    \end{minipage}
\end{table}

\begin{table}[t]
    \centering
        \setlength{\tabcolsep}{4.0mm}
    \renewcommand\arraystretch{1.0}
    \begin{minipage}[t]{0.45\linewidth}
    \scalebox{0.9}{
    \begin{tabular}{c|c|c}
\Xhline{3\arrayrulewidth}
        &   mIoU & AP \\
 \hline
\cellcolor{Graylight}w/ Stop Gradient & \cellcolor{Graylight}\textbf{45.7} & \cellcolor{Graylight}\textbf{43.0} \\
 w/o Stop Gradient & 45.2 & 42.1 \\
     \Xhline{3\arrayrulewidth}

    \end{tabular}
    }
    \vspace{0.1em}
    \caption{Ablation study on the stop gradient technique.}
    \vspace{-1em}
    \label{tab:stop}
    \end{minipage}
    \hspace{3em}
    \begin{minipage}[t]{0.45\linewidth}
        \centering
        \scalebox{0.9}{
        \begin{tabular}{l|cc} %
     \Xhline{3\arrayrulewidth}
        $\epsilon$ &
        mIoU  & AP  \\
        \hline
        0   & 44.9 & 42.5 \\
        0.001   & 45.3 & 42.7  \\
                \cellcolor{Graylight}0.01   & \cellcolor{Graylight}\textbf{45.7} & \cellcolor{Graylight}\textbf{43.0} \\
        0.1   & 44.4 & 41.5 \\
         \Xhline{3\arrayrulewidth}
    \end{tabular}
    }
    \vspace{0.1em}
    \caption{ Different noise level against ADE20k performance. }
    \vspace{-1em}
    \label{tab:noise}
    \end{minipage}
\end{table}

\section{Inference Speed and FLOPs}

\lm{An appealing property for MoE transformer is that it enlarges the model capacity (number of parameter) by only introducing minimal additional inference time and FLOPs than non-MoE transformerts.
RMoE also has this property.
As shown in Table~\ref{tab:inference}, we analyzed the the inference time and FLOPs of Swin-T and Swin-L on ADE20K segmentation tasks using UperNet decoder head.
We show that RMoE has an obvious performance gain but only introduces minimal additional inference time, which is critical in machine learning models deployment.
}

\section{Viable Setting for RMoE}

From the ablation studies, we conclude an effective but simple baseline for practitioners, i.e., the Every-Last strategy for choosing MoE layers, top-1 gate, and 8 experts. According to Table 5 in the main paper, compared with the best performance RMoE-I with firefly splitting, this configuration get the same performance with only 1M additional parameters on the backbone.

\section{Conclusion}
\label{sec:conclusion}
In this paper, we propose an efficient training pipeline for MoE vision transformers, RMoE. 
Compared with traditional MoE training suffering from high computation cost, we train the different residuals for each expert by initializing the weights from the pretrained non-MoE model. This is motivated by analyzing the training trajectory of the weights, where we find that experts weights can be factored as a weight core and a residual. 
We have comprehensive studies on various transformers, including Swin Transformer, CvT and BeiT, on different downstream tasks, including ADE20K segmentation and COCO object detection. We show that our RMoE can improve the performance of regular non-MoE vision transformers with minor additional cost.

\clearpage

\bibliographystyle{splncs04}
\bibliography{egbib}

\newpage
\appendix

\section{Detail for Firefly Splitting pipeline}

Given a trained non-MoE vision transformer, we employ the Firefly approach to select which MLP layers to be replaced by MoE layers.

In particular, it aims at minimizing the loss decrease between non-MoE model $f_{\text{MLP}}$ and the RMoE model $f_{\text{RMoE}}; \bm{\theta_r},  \bm{\theta_g}$:
\begin{equation}
\label{eq:ff}
    \mathop{\text{max}}\limits_{\bm{\theta_r},  \bm{\theta_g}}\{ \mathcal{L}(f_{\text{MLP}}) - \mathcal{L}(f_{\text{RMoE}}; \bm{\theta_r},  \bm{\theta_g}) \},
\end{equation}
where $\bm{\theta_r}$ is the weights residual and $\bm{\theta_g}$ is the gate parameters. 
The optimization problem is solved by gradient descend.
To accelerate the optimization, $\bm{\theta_r}$ is initialized as $\bm{\theta_r} \leftarrow \epsilon\bm{\theta_0}$, where $\epsilon=0.001$ in our experiments.
Such initialization combined with the stop gradient technique ensures the performance will not drop too much initially, e.g., $ \mathcal{L}(f_{\text{MLP}}) = \mathcal{L}(f_{\text{RMoE}}; \bm{\theta_r},  \bm{\theta_g})$ when $\epsilon=0$.

In practice, to save computation cost and avoid overfitting, we do not add MoE layers in all transformer blocks, but select a subset of layers $\{l\}$ with residual weights $\{\theta_r^l\} \subset \bm{\theta_r}$, which contribute most to Eq.\ref{eq:ff}.
To this end, we follow Firefly and use a two-step optimization to find a sparse solution for Eq.\ref{eq:ff}:

\noindent \textit{Step one.} Generate an over-grown MoE model by replacing all MLP layers with MoE layers and optimize $\bm{\theta_r}, \bm{\theta_g}$ using gradient descent for a few steps.

\noindent \textit{Step two.} Fix $\bm{\theta_g}$, re-optimize $\bm{\theta_r}$ by selecting the best candidates ${\theta_r^l}$.
To achieve this, we use Taylor expansion and choose the largest candidates set follows Eq.\ref{eq:firefly} and Eq.\ref{eq:solution}. We summarize the overall RMoE gradient grow based algorithm in Algorithm~\ref{alg:1}.

\begin{algorithm}[t]
\caption{Insert RMoE to vision transformer for downstream task}
\label{alg:1}
\begin{algorithmic}[1]
\STATE \textbf{Input:} a well-trained $L$-layer vision transformer $f$, the loss functions on $m$ given downstream tasks $\mathcal{L}_t(f), t=1...m$, maximum number layer to expand $\mathcal{L}_{\text{max}}$, number of experts to expand $n$. Stage to expand (Intermediate, Downstream).
\STATE Begin before the given stage.
\STATE For $t = 1...m$, calculate the score for each layers using Eq.~\ref{eq:firefly} and sum them together along the tasks as the final score $s_l = \sum_{i=1}^m |s_l|_i$ where $|s_l|_i$ is layer $l$ score for task $i$.
\STATE Select the top-$\mathcal{L}_{\text{max}}$ score, apply RMoE to expand the MLP layer to $n$-expert RMoE at the corresponding layer with the top score.
\STATE Finetune the expanded RMoE model on the downstream tasks.

\end{algorithmic}
\end{algorithm}

\section{Expert Visualization}

To better understand what experts learn in RMoE training, we present the visualization of the patch routing, which represents how each patch in the input token sequence is routed by the gate. 
In an MoE model, each expert is trying to learn different functions to process different kind of input tokens for the input images. 
Thus, we can use this visualization to qualitatively compare the model trained with RMoE and MoE.

\begin{figure*}[ht]
    \centering
    \includegraphics[width=\textwidth]{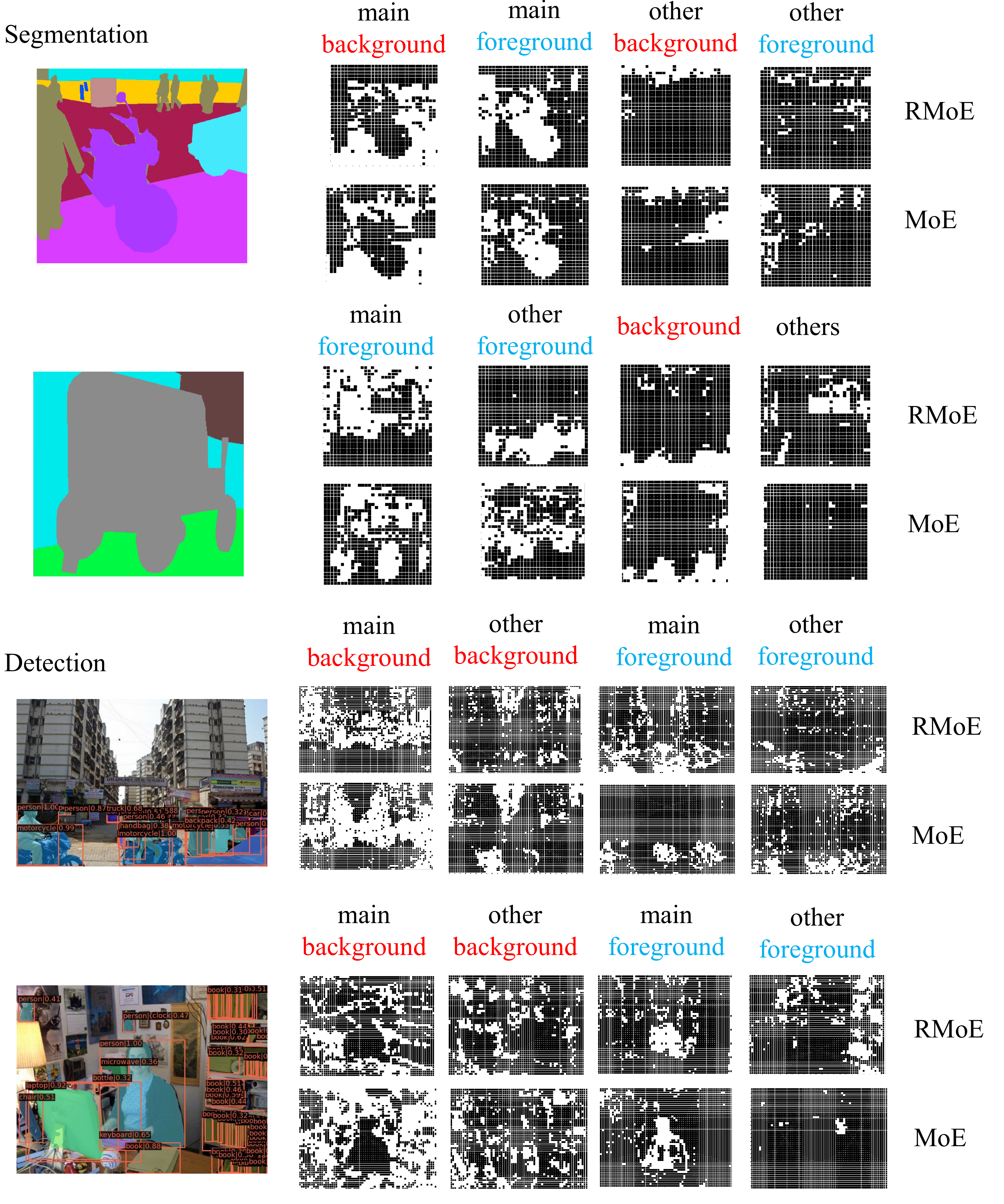}
    \caption{Patch routing visualization for the last layer of Swin-T stage 3 in segmentation and detection tasks. We define the main/other mainly based on number of patches. }
    \label{fig:vis}
\end{figure*}

To visualize the patch routing, we record the patch routing result for each expert and reshape the input patch token sequence to the image-like shape of the current transformer stage. In this way, we can directly visualize the patch routing condition in an image fashion as shown in Figure~\ref{fig:vis}. In the figure, the white square represents the patch is routed to the this expert. The black square means the patch is not routed to this expert. We choose top-4 experts with the most number of routed patches to provide a clearer visualization.

We use Swin-T as backbone and visualize the token routing in the last layer of stage 3. We choose this layer because stage 3 has a proper resolution to visualize, and its feature map contains more semantic information compared with the shallow stages. For segmentation and detection tasks, we use the same setting in Section \ref{sec:exp}. For RMoE, we use RMoE-D models trained on the downstream tasks.

From Figure \ref{fig:vis}, we see that RMoE and MoE share similar patterns in patch routing, for example, the foreground and background are routed into different experts. These indicate the experts of the RMoE share similar functions as the MoE experts.

\section{Additional Implementation detail}
For downstream training, we build our code base on the MMSegmentation~\cite{mmseg2020} and MMDetection~\cite{mmdetection}. For MoE implementation, we use FastMoE~\cite{he2021fastmoe} as basic code.

\end{document}